\documentclass[sigconf]{acmart}

\AtBeginDocument{%
  \providecommand\BibTeX{{%
    \normalfont B\kern-0.5em{\scshape i\kern-0.25em b}\kern-0.8em\TeX}}}

\usepackage{caption}
\usepackage{subcaption}
\usepackage{multirow}
\usepackage{adjustbox}
\usepackage[normalem]{ulem}

\usepackage{algorithm}
\usepackage{algpseudocode}

\definecolor[named]{cool}{RGB}{59, 77, 191}
\definecolor[named]{warm}{RGB}{180, 4, 39}

\newcommand{\systemname}{\textsc{Candle}}
\newcommand{\website}{{\url{https://candle.mpi-inf.mpg.de/}}}

\newcommand{\stmt}[1]{{\sc #1}}

\newcommand{\squishlist}{
 \begin{list}{$\bullet$}
  { \setlength{\itemsep}{0pt}
     \setlength{\parsep}{3pt}
     \setlength{\topsep}{3pt}
     \setlength{\partopsep}{0pt}
     \setlength{\leftmargin}{1.5em}
     \setlength{\labelwidth}{1em}
     \setlength{\labelsep}{0.5em} } }

\newcommand{\squishend}{
  \end{list}  }

\algnewcommand{\algorithmicand}{\textbf{ and }}
\algnewcommand{\algorithmicor}{\textbf{ or }}
\algnewcommand{\OR}{\algorithmicor}
\algnewcommand{\AND}{\algorithmicand}

\renewcommand{\paragraph}[1]{\smallskip\noindent\textbf{#1.\mbox{\ \ }}}

\copyrightyear{2023}
\acmYear{2023}
\setcopyright{rightsretained}
\acmConference[WWW '23]{Proceedings of the ACM Web Conference 2023}{May 1--5, 2023}{Austin, TX, USA}
\acmBooktitle{Proceedings of the ACM Web Conference 2023 (WWW '23), May 1--5, 2023, Austin, TX, USA}
\acmDOI{10.1145/3543507.3583535}
\acmISBN{978-1-4503-9416-1/23/04}

\begin{document}

\title{Extracting Cultural Commonsense Knowledge at Scale}


\author{Tuan-Phong Nguyen}
\affiliation{%
  \institution{Max Planck Institute for Informatics}
  \city{Saarbr\"uken}
  \country{Germany}}
\email{tuanphong@mpi-inf.mpg.de}

\author{Simon Razniewski}
\affiliation{%
  \institution{Max Planck Institute for Informatics}
  \city{Saarbr\"uken}
  \country{Germany}}
\email{srazniew@mpi-inf.mpg.de}

\author{Aparna Varde}
\affiliation{%
  \institution{Montclair State University}
  \city{Montclair}
  \state{New Jersey}
  \country{USA}}
\email{vardea@montclair.edu}

\author{Gerhard Weikum}
\affiliation{%
  \institution{Max Planck Institute for Informatics}
  \city{Saarbr\"uken}
  \country{Germany}}
\email{weikum@mpi-inf.mpg.de}

\begin{abstract}
Structured knowledge is important for many AI applications. 
%
Commonsense knowledge, which is crucial for robust human-centric AI, is covered by a small number of structured knowledge projects. 
However, they lack knowledge about human traits and behaviors conditioned on socio-cultural contexts, which is crucial for situative AI.
This paper presents \systemname{}, an 
end-to-end methodology for extracting high-quality 
cultural commonsense knowledge (CCSK) at scale. 
\systemname{} 
extracts CCSK assertions from a huge web corpus and organizes them into coherent clusters, for 3 domains of subjects (geography, religion, occupation) and 
several cultural facets (food, drinks, clothing, traditions, rituals, behaviors).
\systemname{} includes judicious techniques for classification-based filtering and scoring of interestingness.
%
%
Experimental evaluations show the superiority of the \systemname{} CCSK collection over prior works, and an extrinsic use case demonstrates the benefits of CCSK for the GPT-3 language model.
Code and data can be 
accessed at
\website{}.

\end{abstract}

\maketitle

\section{Introduction}


\paragraph{Motivation}
Structured knowledge, often stored in knowledge graphs (KGs) \cite{hoganetal,weikum2020machine}, is a key asset for many AI applications, including search, question answering, and conversational bots. KGs cover factual knowledge about notable entities such as singers, songs, cities, sports teams, etc. However, even large-scale KGs deployed in practice hardly touch on the dimension of commonsense knowledge (CSK): properties of everyday objects, behaviors of humans, and more. 
Some projects, such as ConceptNet \cite{speer2017conceptnet}, Atomic \cite{atomic}, and Ascent++ \cite{ascentpp} have compiled large sets of CSK assertions, but are solely focused on ``universal CSK'': assertions that are agreed upon by almost all people and are thus viewed as ``globally true''.
What is missing, though, is that CSK must often be viewed in the {\em context of specific social or cultural groups}: the world view of a European teenager does not necessarily agree with those of an American business person or a Far-East-Asian middle-aged factory worker.

\begin{figure}[t]
    \centering
    \includegraphics[width=1.05\columnwidth]{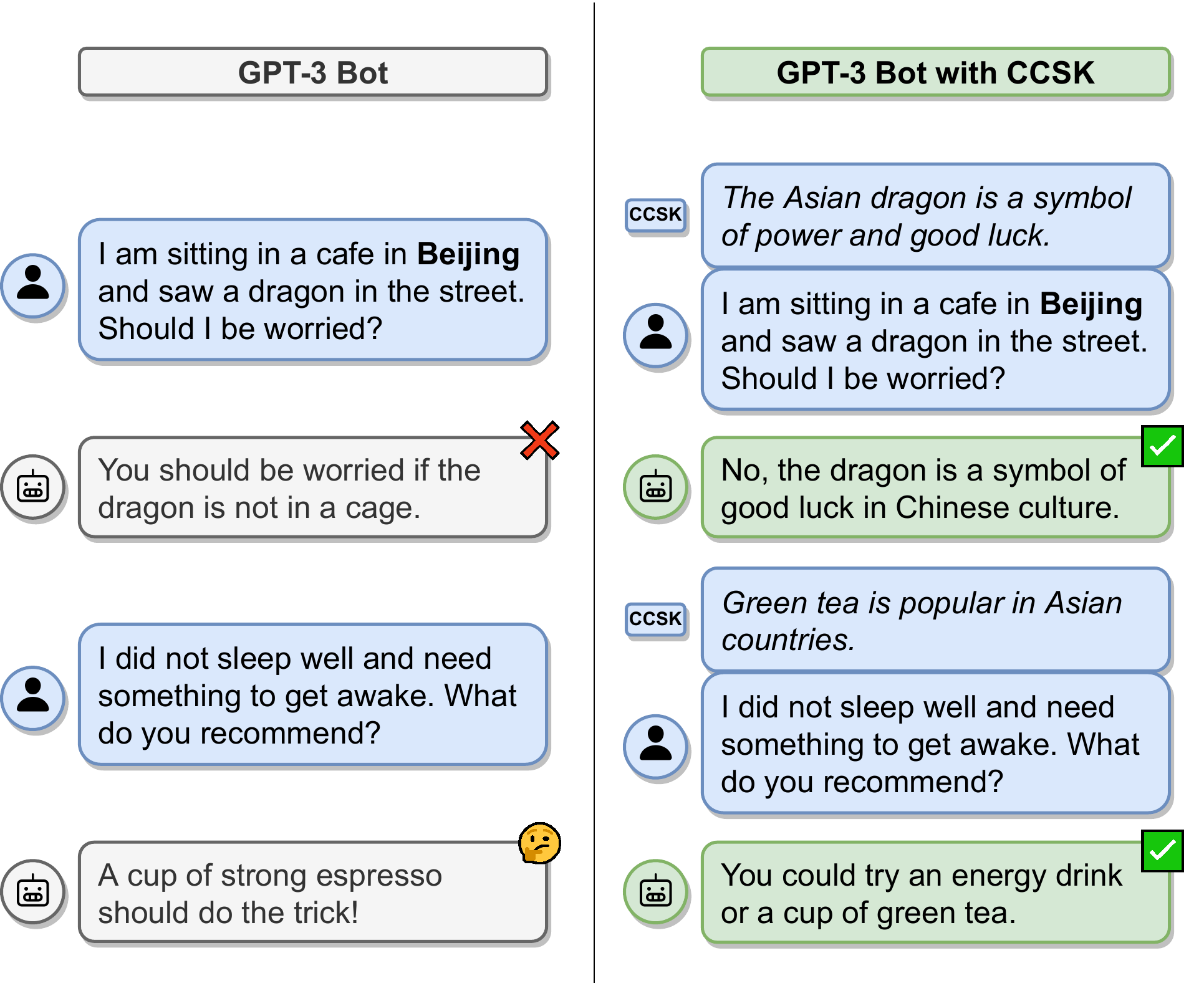}
    \caption{Human-bot conversations without and with CCSK.
    }
    \label{fig:bot-scenario}
\end{figure}

This paper addresses this gap, by automatically compiling CSK that is conditioned on socio-cultural contexts. We refer to this as {\em cultural CSK} or {\em CCSK} for short. 
For example, our CCSK collection contains assertions such as:
\squishlist
\item \stmt
{subject:East Asia, facet:food, Tofu is a
major 
ingredient in 
many 
East Asian cuisines}, or
\item \stmt
{subject:firefighter, facet:behavior, Firefighters use ladders to reach fires}. 
\squishend
The value of having a KG with this information lies in making AI applications more situative and more robust. 

Consider the  conversation between a human and the GPT-3 chatbot\footnote{Executed at \href{https://beta.openai.com/playground}{beta.openai.com/playground} using the \textit{davinci-002} model at temp=0.7.} shown in Fig.~\ref{fig:bot-scenario}.
The GPT-3-based bot, leveraging its huge language model, performs eloquently in this conversation, but completely misses the point that the user is in China, where dragons are viewed positively and espresso is difficult to get.
If we prime the bot with CCSK about Far-East-Asian culture, then GPT-3 is enabled to provide culturally situative replies.
If primed with CCSK about European views (not shown in Fig.~\ref{fig:bot-scenario}), the bot points out that dragons are portrayed as evil monsters but do not exist in reality and recommends a strong cup of coffee.

\paragraph{State of the art}
Mainstream KGs do not cover CCSK at all, and major CSK collections like ConceptNet contain only very few culturally contextualized assertions.
To the best of our knowledge, the only prior works with data that have specifically addressed the socio-cultural dimension are the projects Quasimodo \cite{quasimodo},
StereoKG \cite{stereokg}, and the work of Acharya et al. \cite{acharya2020towards}.
The latter merely contains a few hundred assertions from crowdsourcing, StereoKG uses a specialized way of automatically extracting stereotypes from QA forums and is still small in size, and Quasimodo covers a wide mix of general CSK and a small fraction of culturally relevant assertions.
These are the three baselines to which we compare our results. 

Language models (LMs) such as BERT \cite{devlin2019bert} or GPT-3 \cite{GPT3}
are another form of machine-based CSK, including CCSK, in principle. However, all LM knowledge is in latent form,
captured in learned values of billions of parameters. 
Knowledge cannot be made explicit; we observe it only implicitly
through the LM-based outputs in applications.
The example of Fig.~\ref{fig:bot-scenario}
demonstrates that even large LMs like GPT-3
do not perform well when
cultural
context matters.


\paragraph{Approach}
CCSK is expressed in text form on web pages and social media, but this is often very noisy and difficult to extract. 
We 
devised
an
end-to-end
methodology and system, called \systemname{} (Extracting \underline{C}ultur\underline{a}l Commo\underline{n}sense Knowle\underline{d}ge at Sca\underline{le}),
to automatically extract and systematically organize a large collection of CCSK assertions.
For scale, we tap into the C4 web crawl \cite{t5},
a huge collection of web pages. This provides 
an opportunity to construct a sizable CCSK collection, but also a challenge in terms of scale and noise.

The output of \systemname{} is a set of 1.1M CCSK assertions, organized into 60K coherent clusters. 
The set is organized by 3 domains of interest --
geography, religion, occupation --
with a total of 386 instances, referred to as {\em subjects} (or cultural groups).
Per subject, the assertions cover 5 
{\em facets} of culture: food, drinks, clothing, rituals, traditions (for geography and religion) or behaviors (for occupations). 
In addition, we also annotate each assertion with its salient \textit{concepts}.
Examples for the computed CCSK are shown in 
Fig.~\ref{fig:socco-examples}.

\systemname{} operates in 6 steps.
First and second, we identify candidate assertions using simple techniques for \textit{subject detection} (named entity recognition - NER, and string matching), and \textit{generic rule-based filtering}.
Third, we \textit{classify assertions} into specific cultural facets, which is challenging because we have several combinations of cultural groups and cultural facets, making it very expensive to create specialized training data.
Instead, we creatively leverage LMs pre-trained on the Natural Language Inference (NLI) task to perform zero-shot classification on our data, with judicious techniques to enhance the accuracy.
Fourth we use state-of-the-art techniques for \textit{assertion clustering}, and fifth a simple but effective method to \textit{extract concepts} in assertions.
Lastly, we combine several features to \textit{score} the interestingness of assertions, such as frequency, specificity, distinctiveness.
This way, we 
steer away from overly generic assertions
(which LMs like GPT-3 tend to generate) and favor assertions that set their subjects apart from others.



\begin{figure}[t]
    \centering
    \small
    \begin{tabular}{|lll|}
        \hline
        \rowcolor{gray!10}
         \textcolor{warm}{\textbf{geography>country}} & \textcolor{Cerulean}{\textbf{Germany}} & \textcolor{LimeGreen}{\textbf{drinks}} \\
         \hline
         \multicolumn{3}{|p{0.95\columnwidth}|}{German \textcolor{YellowOrange}{\textbf{beer festivals}} in October are a celebration of beer drinking.} \\
         \hline
        \rowcolor{gray!10}
         \textcolor{warm}{\textbf{geography>region}} & \textcolor{Cerulean}{\textbf{East Asia}} & \textcolor{LimeGreen}{\textbf{food}} \\
         \hline
         \multicolumn{3}{|p{0.95\columnwidth}|}{\textcolor{YellowOrange}{\textbf{Tofu}} is a major ingredient in many East Asian cuisines.} \\
         \hline
        \rowcolor{gray!10}
         \textcolor{warm}{\textbf{geography>region}} & \textcolor{Cerulean}{\textbf{South Asia}} & \textcolor{LimeGreen}{\textbf{traditions}} \\
         \hline
         \multicolumn{3}{|p{0.95\columnwidth}|}{In South Asia, \textcolor{YellowOrange}{\textbf{henna}} is often used in bridal makeup or to celebrate festivals.} \\
         \hline
        \rowcolor{gray!10}
         \textcolor{warm}{\textbf{occupation}} & \textcolor{Cerulean}{\textbf{lawyer}} &
         \textcolor{LimeGreen}{\textbf{clothing}}
         \\
         \hline
         \multicolumn{3}{|p{0.95\columnwidth}|}{Lawyers wear \textcolor{YellowOrange}{\textbf{suits}} to look professional.} \\
         \hline
        \rowcolor{gray!10}
         \textcolor{warm}{\textbf{occupation}} & \textcolor{Cerulean}{\textbf{firefighter}} & 
         \textcolor{LimeGreen}{\textbf{behaviors}} 
         \\
         \hline
         \multicolumn{3}{|p{0.95\columnwidth}|}{Firefighters run into burning buildings to \textcolor{YellowOrange}{\textbf{save lives}}.} \\
         \hline
    \end{tabular}

    \caption{Example assertions of \systemname{}, with \textcolor{Cerulean}{\textbf{subjects}} (cultural groups) of cultural \textcolor{warm}{\textbf{domains}},  \textcolor{LimeGreen}{\textbf{facets}} and \textcolor{YellowOrange}{\textbf{concepts}}.
    }
    \label{fig:socco-examples}
\end{figure}

\paragraph{Contributions}
The main contributions of this work are:
\begin{enumerate}
    \item An 
    end-to-end methodology to extract high-quality CCSK from very large text corpora. 
    \item New techniques for judiciously classifying and filtering CCSK-relevant text snippets, and for
    scoring assertions by their interestingness. 
    \item A large collection of CCSK assertions for 386 subjects covering 3 domains (geography, religion, occupation) and several facets (food, drinks, clothing, traditions, rituals, behaviors).
\end{enumerate}

Experimental evaluations show that the assertions in \systemname{} are of significantly higher quality than those from prior works.
An extrinsic use case demonstrates that our CCSK can improve performance of GPT-3 in question answering.
%
Code and data can be 
accessed at \website{}.

\section{Related work}
\paragraph{Commonsense knowledge acquisition}
There is a long tradition of CSK acquisition in AI (e.g., \cite{lenat1995cyc,singh2002open,liu2004conceptnet,DBLP:conf/aaai/GordonDS10,10.1145/3437963.3441664}).
Earlier projects, e.g., Cyc~\cite{lenat1995cyc} and ConceptNet~\cite{liu2004conceptnet}, construct commonsense knowledge graphs (CSKGs) based on large-scale human annotations.
Crowdsourcing CSKG construction has been revived in the ATOMIC project~\cite{atomic,comet-atomic-2020}. CSK extraction from texts has been researched in WebChild~\cite{webchild}, TupleKB~\cite{tuplekb}, Quasimodo~\cite{quasimodo}, ASER~\cite{zhang2020aser,ZHANG2022103740}, TransOMCS~\cite{transomcs}, GenericsKB~\cite{bhakthavatsalam2020genericskb}, and Ascent~\cite{ascent,ascentpp}.
Meanwhile, Ilievski et al.~\cite{cskg}
consolidate CSK from 7 different resources into one integrated KG.
Those projects, however, have their main focus on either concept-centered knowledge (e.g., \stmt{Elephants have trunks}), social interactions (e.g., {\sc X hates Y's guts, as a result, X wants to yell at Y}), or event-centered knowledge (e.g., \stmt{X drinking coffee happens after X pouring the coffee into a mug}) and do not cover much cultural knowledge.
Our approach also starts from texts, but focuses on 
cultural commonsense
knowledge (CCSK), with particular challenges in knowledge representation, assertion filtering and consolidation.

\paragraph{Cultural commonsense knowledge}
A few works have focused specifically on CCSK.
An early approach by Anacleto et al.~\cite{Anacleto2006} gathers CSK from users from different cultures, entered via the Open Mind Common Sense portal. However, the work is limited to a few eating habits (time for meals, what do people eat in each meal?, food for party/Christmas) in 3 countries (Brazil, Mexico, USA), and without published data.
Acharya et al.\ \cite{acharya2020towards} embark on a similar manual effort towards building a cultural CSKG, limited to a few predefined predicates and answers from Amazon MTurk workers from USA and India.
Shwartz \cite{shwartz-2022-good} maps time expressions in 27 different languages to specific hours in the day, also using MTurk annotations. 
StereoKG~\cite{stereokg} mines cultural stereotypes of 5 nationalities and 5 religion groups from Twitter and Reddit questions posted by their users, however, being without proper filtering, the method results in quite many noisy and inappropriate assertions. 
GeoMLAMA~\cite{geomlama} defines 16 geo-diverse commonsense concepts (e.g., traffic rules, date formats, shower time) and use crowdsourcing to collect knowledge for 5 different countries in 5 corresponding languages. The dataset was used to probe multilingual pretrained language models, however, is not shared.
Moving to computer vision, Liu et al. \cite{liu-etal-2021-visually} and Yin et al. \cite{yin-etal-2021-broaden} expand existing visual question answering datasets with images from different cultures rather than the Western world. 
As a result, models trained on images from the old datasets (mostly images from Western cultures) perform poorly on the newly added images. 
Our methodology is the first to utilize large text corpora, and it can extract CCSK in the form of natural-language sentences, for a wide range of cultural groups and facets.


\paragraph{Pre-trained language models and commonsense knowledge}
Remarkable advances in NLP have been achieved with pre-trained language models (LMs) such as BERT~\cite{devlin2019bert} and GPT variants~\cite{radford2019language,GPT3}.
LAMA~\cite{petroni2019language} designs methodology and datasets to probe masked LMs in order to acquire CSK that the models implicitly store.
COMET~\cite{bosselut2019comet} is a method that finetunes autoregressive LMs on CSK triples, and it can generate possible objects for a given pair of subject-predicate. However, the quality of the generated assertions is often considerably lower than that of the training data~\cite{nguyen-razniewski-2022-materialized}. 
More recently, West et al. \cite{west2021symbolic} introduce a prompting technique to collect CSK by feeding GPT-3~\cite{GPT3} with a few human-verified CSK triples and ask it to generate new assertions. Although it was shown that the generated resource
is of encouraging quality, knowledge bases from LMs are inherently problematic, because their is no apparent way to trace assertions to specific sources, e.g., to understand assertion context, or to apply filters at document level.

In this work, we leverage pre-trained LMs as sub-modules in our system to help with cultural facet classification and assertion clustering.
We also show that our method can produce more 
distinctive CCSK assertions than querying GPT-3 with 
prompts.



\section{CCSK Representation}

Our representation of CCSK is based on the notions of {\em subjects} (from 3 major domains: geography, religion and occupation) and {\em facets}.
These are the key labels for CCSK {\em assertions}, which are informative sentences with salient {\em concepts}
marked up.

We assume two sets to be given:
\begin{itemize}
    \item $\mathcal{S}$: A set of \textbf{subjects} (cultural groups)
    $s_1, \ldots, s_n$ from a cultural \textbf{domain}, e.g., based on geo-locations (United States, China, Middle East, California), religious groups (Christians, Muslims, Buddhists) or occupations (taxi driver, professor, web developer);
    \item $\mathcal{F}$: A set $F_1, \ldots, F_m$ of \textbf{facets} of culture, e.g., food, drinks, clothing, traditions, rituals, behaviors.
\end{itemize}
Note that the cultural facets need not be mutually exclusive, e.g., food assertions sometimes overlap with traditions.

Our objective is to collect a set of CCSK assertions for a given subject and facet.
Existing commonsense resources store assertions in triple format (e.g., ConceptNet~\cite{speer2017conceptnet}, 
Quasimodo~\cite{quasimodo}), semantic frames (Ascent~\cite{ascent}) or generic sentences (GenericsKB~\cite{bhakthavatsalam2020genericskb}).
Although the traditional triple-based and frame-based data models are convenient for structured querying, and well suited for regular assertions like birth dates, citizenships, etc., they often falls short of capturing nuanced natural-language assertions, as essential for CSK. Moreover, recent advances in pre-trained LMs have made it easier to feed downstream tasks with 
less structured knowledge.

With \systemname{}, we thus follow the approach of GenericsKB \cite{bhakthavatsalam2020genericskb}, and use natural-language sentences to represent assertions.

In principle, an assertion could comprise even several sentences. The longer the assertions are, however, the harder it is to discern their core. In this work, for higher precision and simplicity of computations, we only consider single sentences.


\newtheorem{definition*}{Definition}

\begin{definition*}[Cultural commonsense knowledge assertion]
Given a subject $s$ and a facet $F$, a CCSK assertion is a triple $(s,F,\textit{sent})$ where \textit{sent} is a  natural-language sentence
about facet $F$ of subject $s$.
\end{definition*}


Since natural language often allows to express similar assertions in many different ways, and web harvesting naturally leads to discovering similar assertions multiple times, we employ clustering as an essential component in our approach.

A \textbf{cluster} (\textit{cls}) of CCSK assertions for one subject and cultural facet contains assertions with similar meaning, and for presentation purposes, is summarized by a single summary sentence. Each cluster also comes with a score denoting its interestingness.

To further organize assertions, we also identify salient \textbf{concepts}, i.e., important terms inside assertions, that can be used for concept-centric browsing of assertion sets.

Several examples of CCSK assertions produced by \systemname{} are shown in Fig.~\ref{fig:socco-examples}.


\section{Methodology}

\begin{figure*}[t]
    \centering
    \includegraphics[width=.9\textwidth]{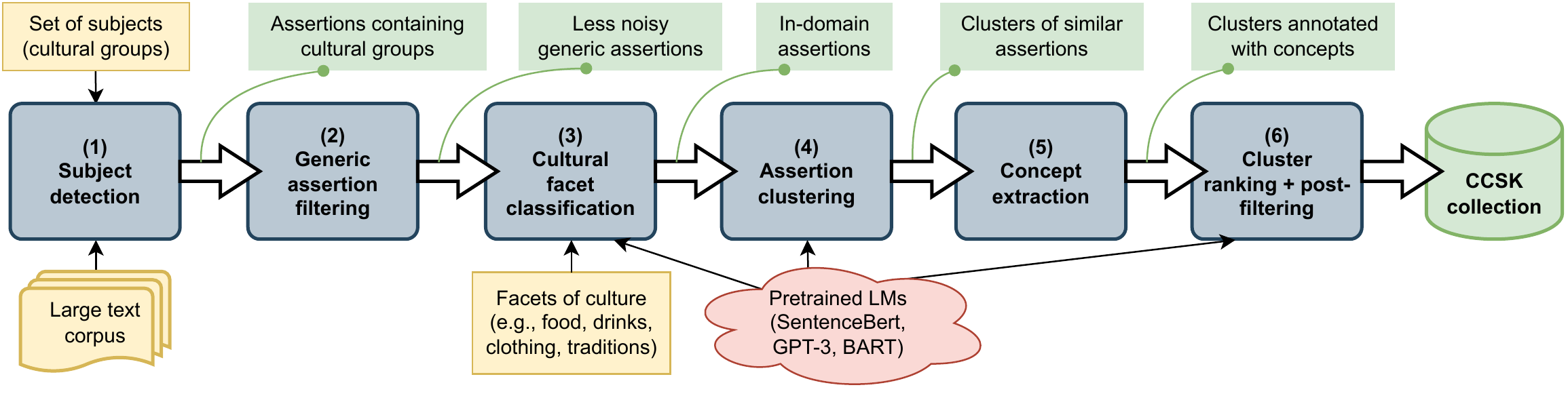}
    \caption{Architecture of \systemname{}
    (see Table~\ref{tab:list-techniques} for the list of techniques and models used in each module).
    }
    \label{fig:architecture}
\end{figure*}

We propose an end-to-end system, called \systemname{}, to extract and organize CCSK assertions based on the proposed CCSK representation.
Notably, our system does not require annotating new training data, but only leverages pre-trained models with judicious techniques to enhance the accuracy. 
The system takes in three inputs:
\begin{itemize}
    \item an English text corpus (e.g., a large web crawl);
    \item a set of \emph{subjects} (cultural groups);
    \item a set of \emph{facets} of culture.
\end{itemize}

\systemname{} consists of 6 modules (see Fig.~\ref{fig:architecture}). Throughout the system, step by step, we reduce a large input corpus (which could contain billions of documents, mostly noisy) into high-quality clusters of CCSK assertions for the given subjects and facets.
Each cluster in the output is also accompanied by a representative sentence and an interestingness score. We next elaborate on each module.

\subsection{Subject detection}\label{sec:subject-detection}

We start the extraction by searching for sentences that contain mentions of the given subjects. These will be the candidate sentences used in the subsequent modules. To achieve high recall, we utilize generous approaches such as string matching and named entity recognition (NER), and use more advanced filtering techniques in later modules, to ensure high precision.



For the geography and religion domains, in which subjects are named entities, we use spaCy's NER module to detect subjects. Specifically, geo-locations are detected with the GPE tag (geopolitical entities), and religions are detected with the NORP tag (nationalities or religious or political groups). For each subject, we also utilize a list of aliases for string matching, which can be the location's alternate names (e.g., United States, the U.S., the States), or demonyms (e.g., Colombians, Chinese, New Yorker), or names for religious adherents (e.g., Christians, Buddhists, Muslims) - which can be detected with the NORP tag as well.

For the occupation domain, we simply use exact-phrase matching to detect candidates. Each occupation subject is enriched with its alternate names and its plural form to enhance coverage.

\subsection{Generic assertion filtering}\label{sec:generic-filtering}

CSK aims at covering \emph{generic assertions}, not episodic or personal experiences. For example, \stmt{Germans like their currywurst} is a generic assertion, but \stmt{I visited Germany to eat currywurst} or \stmt{This restaurant serves German currywurst} are not.

GenericsKB~\cite{bhakthavatsalam2020genericskb} is arguably the most popular work on automatically identifying generic sentences in texts and it uses a set of 27 hand-crafted lexico-syntactic rules.
\systemname{} adopts those rules in this module. 
However, for each domain and facet, we adaptively drop some of the rules if they would reject valuable assertions. 
More details on the adaptations can be found in Appx.~\ref{apd:filtering}.

\subsection{Cultural facet classification}\label{sec:classification}
To organize CCSK and filter out irrelevant assertions, we classify candidate sentences into several facets of culture.
%
Traditional methods for this classification task would require a substantial amount of annotated data to train a supervised model.
The costs of data annotation are 
often a critical bottleneck in large-scale settings.
In \systemname{}, we aim to minimize the degree of human supervision by leveraging
pre-trained models for zero-shot classification.

A family of pre-trained models that is suitable for our setting is {\em textual entailment} (a.k.a {\em natural language inference} - NLI): given two sentences, does one entail the other (or are they contradictory or unrelated)?
Our approach to adopting such a model for
cultural facet classification is inspired by
the zero-shot inference method of
Yin et al.~\cite{yin-etal-2019-benchmarking}. 
%
Given a sentence $sent$ and a facet $F$, we construct the NLI test as follows: 
%
\begin{algorithm}[htbp]
\begin{algorithmic}
\Require $\textit{Premise} \gets \textit{sent}, \textit{Hypothesis} \gets \text{``This text is about } F \text{''}$
\Ensure $P[\textit{sent} \in F] \gets P[\textit{Premise} \Rightarrow \textit{Hypothesis}]$
\end{algorithmic}
\end{algorithm}
%

The probability of $Premise$ entailing $Hypothesis$ will be taken as the probability of $sent$ being labeled as $F$, denoted as $P[sent \in F]$.
For example, with sentence ``German October festivals are a celebration of beer and fun'',
the candidate entailments will be ``This text is about drinks'', ``... about food'', ``... about traditions'', and so on. 
Multiple of these facets may yield high scores in these NLI tests.

To enhance precision, we introduce a set of \textit{counter-labels} for topics that are
completely outside the scope of CCSK, for
example, politics or business.
A sentence $sent$ will be accepted as a good candidate for facet $F$ if
\begin{equation}\label{eq:classification}
\begin{cases}
P[sent \in F] \ge \rho_+ \text{, and} \\
    P[sent \in \tilde{F}] \le \rho_- \text{ for all counter-labels } \tilde{F}
\end{cases}
\end{equation}
where $\rho_+$ and $\rho_-$ are hyperparameters in the range $[0, 1]$, giving us the flexibility to tune for either precision or recall.

In our experiments, we use the BART model~\cite{lewis2019bart} finetuned on the MultiNLI dataset~\cite{williams-etal-2018-broad} for NLI tests\footnote{Model available at \url{https://huggingface.co/facebook/bart-large-mnli}}.
%
Our crowdsourcing evaluations show that the zero-shot classifiers with the enhanced techniques achieved high precision (see Appx.~\ref{subsec:per-facet-quality}).


\subsection{Assertion clustering}\label{sec:clustering}
The same assertion can be expressed in many ways in natural language.
For example, \stmt{Fried rice is a popular Chinese dish} can also be written as \stmt{Fried rice is a famous dish from China} or \stmt{One of the most popular Chinese food is fried rice}.
Clustering is used to group such assertions, which reduces redundancies, and allows to obtain frequency signals on assertions.


We leverage a state-of-the-art sentence embeddings method, SentenceBert~\cite{sbert}, to compute vector representations for all assertions and use the Hierarchical Agglomerative Clustering (HAC) algorithm for clustering.
Clustering is performed on assertions of each subject-facet pair. 


\paragraph{Cluster summarization}
Since each cluster can have from a few to hundreds of sentences, it is important to identify what those sentences convey, in a concise way.

One 
way to 
compute 
a representative assertion for a cluster is to compute the centroid of the cluster,
then take its closest assertion as the representative.
Yet for natural-language data, this does not work particularly well.

In \systemname{}, we therefore approach cluster summarization as a generative task, based on a state-of-the-art 
LM,
GPT-3~\cite{GPT3} (see Appx.~\ref{apd:gpt-prompt} for prompt template).
% ...
%
Annotator-based evaluations show that GPT-generated representatives received significantly better scores than the base sentences in the clusters (see Sec.~\ref{sec:comparison-other-resources}).

\subsection{Concept extraction}\label{sec:concept-extractino}
While the cultural groups are regarded as subjects, \textit{concepts} are akin to objects of the assertions. Identifying these concepts enables concept-focused browsing (e.g., browsing Japan assertions only about the Miso soup, etc.).

We postulate that main concepts of an assertion cluster are terms shared by many members: We extract all n-grams ($n=1..3$) of all assertions in a cluster (excluding subjects themselves, and stop words); and retain the ones that occur in more than 60\% of the assertions. If both a phrase and its sub-phrase appear, we only keep the longer phrase in the final output. 
Noun-phrase concepts are normalized by singularization.

\subsection{Cluster ranking and post-filtering}\label{sec:ranking}

Ranking commonsense assertions is a crucial task.
Unlike encyclopedic knowledge, which is normally either true or false, precision of CSK is usually not a binary concept, as it generalizes over many groups.
With \systemname{}, we aim to pull out the most interesting assertions for each subject, and avoid overly generic assertions such as \stmt{Chinese food is good} or \stmt{Firefighters work hard}, which are very common in the texts.

Extracting and clustering assertions from large corpora gives us an important signal of an assertion, its \textit{frequency}.
However, ranking based on frequency alone may lead to reporting bias.
As we compile a CCSK collection at large scale, it also enables us to compute the \textit{distinctiveness} of an assertion against others in the collection.
The notion of these 2 metrics can be thought of as term frequency and inverse document frequency in the established TF-IDF technique for IR document ranking \cite{10.5555/106765.106782}.
Besides \textit{frequency} and \textit{distinctiveness}, we score the interestingness of assertion clusters based on 2 other custom metrics: \textit{specificity} (how many objects are mentioned in the assertion?) and \textit{domain relevance} (how relevant is the assertion to the cultural facet?).


\paragraph{Frequency}
For each subject-facet pair, we normalize cluster sizes into the range $[0,1]$, using min-max normalization.

\paragraph{Distinctiveness}
We compute the IDF of a cluster $cls$ as follows:
\begin{equation}
    IDF(cls) = \frac{\sum_{cls' \in CLS}size(cls')}{\sum_{cls' \in CLS}size(cls') \times \sigma(cls,cls')}
\end{equation}
where $CLS$ is the set of all clusters for a given facet (e.g., food) and domain (e.g., geography>country), and
\begin{equation}\label{eq:distinctiveness}
    \sigma(cls,cls') = 
    \begin{cases}
    1 & \quad \text{if } sim(cls, cls') \ge \theta\\
    0 & \quad \text{otherwise}
    \end{cases}
\end{equation}
Here, $sim(cls, cls')$ is the semantic similarity between the two clusters $cls$ and $cls'$, and $\theta$ is a predefined threshold.
In \systemname{}, to reduce computation, we approximate $sim(cls, cls')$ as the similarity between their summary sentences, which can be computed as the cosine similarity between their embedding vectors.
When computing these embeddings, the subjects in the sentences are replaced with the same [MASK] tokens so that we only compare the expressed properties.
Then, we normalize the logarithmic IDF values into the range $[0, 1]$ to get the distinctiveness scores of clusters.

\paragraph{Specificity}
We compute the specificity of an assertion based on the fraction of nouns in it.
Concretely, in \systemname{}, the specificity of a cluster is computed as the specificity of its summary sentence.

\paragraph{Domain relevance} For each facet, we compute the domain relevance of a cluster by taking the average of the probability scores given to its members by the cultural facet classifier.

\paragraph{Combined score}
The final 
score for cluster $cls$ is the average of the 4 feature scores. A higher score means higher interestingness.

\paragraph{Post-filtering}
Lastly, to eliminate redundancies
and noise, 
and further improve the final output quality, we employ a few hand-crafted rules:
\begin{itemize}
    \item At most 500 clusters per subject-facet pair are retained, as further clusters mostly represent redundancies or noise.
    \item We remove clusters that have no concepts extracted, or that are based on too few distinct sentences ($>$2/3 same sentences) or web source domains.
    \item We remove any cluster if either its summary sentence or many of its member sentences match a bad pattern. We compile a set of about 200 regular expression patterns,
    which were written by a knowledge engineer in one day. 
    For e.g., we reject assertions that contain ``the menu'', ``the restaurant'' (likely advertisements for specific restaurants), or animal and plant breeds named after locations, such as ``American bison'', ``German Shepherd'', etc.
\end{itemize}


\section{Implementation}

\begin{table}[t]
\centering
\caption{Statistics of the \systemname{} CCSK collection
(\#A: number of assertions, \#C: number of clusters).
}
\label{tab:statistics}
\small
\begin{tabular}{lrrrrrr}
\toprule
\multirow{2}{*}{\textbf{Facet}} & \multicolumn{2}{c}{\textbf{Geography}} & \multicolumn{2}{c}{\textbf{Religions}} & \multicolumn{2}{c}{\textbf{Occupations}} \\
 \cmidrule(lr){2-3} \cmidrule(lr){4-5} \cmidrule(lr){6-7}
 & \textbf{\#A} & \textbf{\#C} & \textbf{\#A} & \textbf{\#C} & \textbf{\#A} & \textbf{\#C} \\
\midrule
Food & 240,459 & 12,981 & 9,750 & 680 & 9,837 & 511 \\
Drinks & 95,394 & 5,923 & 3,079 & 218 & 3,321 & 227 \\
Clothing & 14,170 & 1,237 & 1,695 & 141 & 4,367 & 278 \\
Rituals & 116,839 & 8,007 & 74,651 & 3,026 & 22,581 & 1,253 \\
Traditions & 214,931 & 13,606 & 68,202 & 2,798 & - & - \\
Behaviors & - & - & - & - & 25,152 & 1,495 \\
Other & - & - & 60,483 & 2,292 & 159,239 & 5,461 \\
\midrule
\rowcolor{gray!25}
\textbf{All} & 681,793 & 41,754 & 217,860 & 9,155 & 224,497 & 9,225 \\
\bottomrule
\end{tabular}
\end{table}

\paragraph{Input corpus}
In \systemname{}, we use the broad web as knowledge source, because of its diversity and coverage, which are important for long-tail subjects. Besides the benefits, the most challenging problem when processing web contents is the tremendous amount of noise, offensive materials, incorrect information etc., hence, choosing a corpus that has been chiefly cleaned is beneficial.
We choose the Colossal Clean Crawled Corpus (C4)~\cite{t5} as our input, a cleaned version of the Common Crawl corpus, created by applying filters such as deduplication, English-language text detection, removing pages containing source code, offensive language, too little content, etc. We use the \textsc{C4.En} split, which contains 365M English articles, each with text content and source URL.
Before passing it to our system, we preprocessed all C4 documents using spaCy, which took 2 days on our cluster of 6K CPU cores.

\paragraph{Subjects}
We collect CCSK for subjects from 3 cultural domains: geography (272 subjects), religions (14 subjects) and occupations (100 subjects). For geography, we split into 4 sub-domains: countries, continents, geopolitical regions (e.g., Middle East, Southeast Asia, etc.) and US states, which were collected from the GeoNames database\footnote{\url{http://www.geonames.org/}}, which also provides alias names. We further enriched these aliases with demonyms from Wikipedia\footnote{\url{https://en.wikipedia.org/wiki/Demonym}}.

\paragraph{Facets of culture}
We consider 5 facets: \textit{food}, \textit{drinks}, \textit{clothing}, \textit{rituals}, and \textit{traditions} (for geography/religion) or \textit{behaviors} (for occupation), selected based on an article on facets of culture \cite{wikipedia-culture}.

\paragraph{Execution and result statistics}
After tuning the system's hyperparameters on small withheld data (see Appx.~\ref{apd:hyperparameters}), we executed \systemname{} on a cluster of 6K CPU cores (AMD EPYC 7702) and 40 GPUs (a mix of NVIDIA RTX 8000, Tesla A100 and A40 GPUs). 

Regarding processing time, for the domain country (196 subjects), it took a total of 12 hours to complete the extraction, resulting in 8.4K clusters for the facet food (cf. Table~\ref{tab:time-and-size}). Occupations and religions took 8 and 6 hours each.

We provide statistics of the output in Table~\ref{tab:statistics}. In total, the resulting collection has 1.1M CCSK assertions (i.e., base sentences) which form 60K clusters for the given subjects and facets.

\begin{table}[t]
    \centering
    \caption{Processing time and output size of each step in \systemname{} for the domain 
    geography>country
    and facet 
    food.
    }
    \label{tab:time-and-size}
    \small
    \begin{tabular}{crcl}
        \toprule
        \textbf{\#} & \textbf{Step} & \textbf{Time} & \textbf{Output/Data size} \\
        \midrule
        & Input & {\begin{tabular}[c]{@{}c@{}}2 days \\ for NLP \\ preprocessing\end{tabular}} & {\begin{tabular}[c]{@{}l@{}}C4 corpus: 8B sentences \\ 
        196 countries \\ 705 alternate names\end{tabular}} \\
        \midrule
        1 & Subject detection & 2 hours & {\begin{tabular}[c]{@{}l@{}}
        367M subject matches \\ 300M sentences (-96\%) \end{tabular}} \\
        \midrule
        2 & {\begin{tabular}[c]{@{}r@{}}Generic assertion \\  filtering\end{tabular}} & 2 hours & {\begin{tabular}[c]{@{}l@{}}
        13M generic sentences \\ (-96\%)\end{tabular}} \\
        \midrule
        3 & {\begin{tabular}[c]{@{}r@{}}Cultural facet \\  classification\end{tabular}} & 4 hours & {\begin{tabular}[c]{@{}l@{}}769K  positive sentences \\ (-94\%) \\ \end{tabular}} \\
        \midrule
        4 & Assertion clustering & 4 hours & {\begin{tabular}[c]{@{}l@{}}42K clusters (-93\%)\end{tabular}}  \\
        \midrule
        5 & Concept extraction & < 5 minutes & 12.4K concepts \\
        \midrule
        6 & {\begin{tabular}[c]{@{}r@{}}Cluster ranking \\ and post-filtering
        \end{tabular}} & < 5 minutes &  8.8K clusters (-80\%) \\
        \midrule
        - & \textbf{Total} & \textbf{\textasciitilde{} 12 hours} & \\
        \bottomrule
    \end{tabular}
\end{table}




\section{Evaluation}


We perform the following evaluations:
\begin{enumerate}
    \item A comparison of quality of \systemname{}'s output and existing socio-cultural CSK resources: This analysis will show that our CCSK collection is of significantly higher quality than existing resources (Sec.~\ref{sec:comparison-other-resources}), and even outperforms GPT-3-generated assertions (Sec.~\ref{sec:comparison-lm}).
    \item Two extrinsic use cases for CCSK: In this evaluation, we perform two downstream applications, question answering (QA) and a ``guess the subject'' game, showing that using CCSK assertions from \systemname{} is beneficial for these tasks, and that \systemname{} assertions outperform those generated by GPT-3 (Sec.~\ref{sec:extrinsic}).
\end{enumerate}
In Appx.~\ref{apd:intrinsic}, we also break down our CCSK collection into domains and facets, analyzing in details the assertion quality for each sub-collection.

\begin{table*}[t]
\centering
\caption{\systemname{} in comparison to other CSK resources. Quality evaluated on assertions of 5 popular countries in StereoKG.
Abbrv.: PLA - plausibility, COM - commonality, DIS - distinctiveness, OFF - offensiveness, LEN - average assertion length. 
}
\label{tab:kb-comparison}
\small
\begin{tabular}{@{}lllrrrrrrrrrr@{}}
\toprule
\multirow{2}{*}{\textbf{Resource}} & \multirow{2}{*}{\textbf{Construction}} & \multirow{2}{*}{\textbf{Format}} & \multicolumn{3}{c}{\textbf{Size}} & \multicolumn{3}{c}{\textbf{Quality [0..2]}} & \multirow{2}{*}{{\begin{tabular}[c]{@{}r@{}}\textbf{OFF} \\ \textbf{(\%)}\end{tabular}}} & \multirow{2}{*}{\textbf{LEN}} \\
\cmidrule(lr){4-6} \cmidrule(lr){7-9}
 &  &  & \textbf{196 countries} & \textbf{10 religions} & \textbf{100 occupations} & \textbf{PLA} & \textbf{COM} & \textbf{DIS}\\ \midrule
\textbf{Acharya et al.}~\cite{acharya2020towards} & Crowdsourcing & Fixed relations &  225 &	0 &	0 &	1.32 &	\textbf{1.22} &	0.25 & 2 & 102 \\
\textbf{StereoKG}~\cite{stereokg} & Text extraction & OpenIE triples &  2,181 &	1,810 &	0 &	0.54 &	0.46 &	0.21 & 18 & 37 \\
\textbf{Quasimodo}~\cite{quasimodo} & Text extraction & OpenIE triples &  22,588 &	10,628 &	51,124 &	0.68 &	0.65 &	0.31  & 13 & 32 \\
\rowcolor{gray!25}
\textbf{\systemname{}-base-sent} & Text extraction & Sentences &  520,971 &	226,807 &	238,057 &	1.21 &	0.93 &	0.76 & 1 & 69 \\
\rowcolor{gray!25}
\textbf{\systemname{}-cluster-reps} & Text extraction & Sentences &  28,711 &	8,823 &	9,826 &	\textbf{1.50} &	1.15 &	\textbf{1.03} & 1 & 73 \\
\bottomrule
\end{tabular}
\end{table*}

\subsection{Comparison with other resources}\label{sec:comparison-other-resources}

\subsubsection{Evaluation metrics}\label{subsec:metrics}
Following previous works~\cite{quasimodo,stereokg}, we analyze assertion quality along several complementary metrics, annotated by Amazon MTurk (AMT) crowdsourcing.
\begin{enumerate}
    \item \textbf{Plausibility (PLA).} This dimension measures whether assertions are considered to be generally true, a CCSK-softened variant of correctness/precision. 
    \item \textbf{Commonality (COM).} This dimension measures whether annotators have heard of the assertion before, as a signal for whether assertions cover mainstream or fringe knowledge (akin to salience).
    \item \textbf{Distinctiveness (DIS).} This dimension measures discriminative informativeness of assertions, i.e., whether the assertion differentiates  the subject from others.
\end{enumerate}
Each metric is evaluated on a 3-point Likert scale for negation (0), ambiguity (1) and affirmation (2). Distinctiveness (DIS) is only applicable if the answer to the plausibility (PLA) question is either 1 or 2.
In case the annotators are not familiar with the assertion, we encourage them to perform a quick search on the web to find out the answers for the PLA and DIS questions.
More details on the AMT setup can be found in Appx.~\ref{apd:human-eval}.

\subsubsection{Compared resources}
We compare \systemname{} with 3 prominent CSK resources: Quasimodo~\cite{quasimodo}, Acharya et al.~\cite{acharya2020towards}, StereoKG~\cite{stereokg}.
The former covers broad domains including assertions for countries and religions, while the others focus on cultural knowledge.
%
Other popular resources such as ConceptNet~\cite{speer2017conceptnet}, GenericsKB~\cite{bhakthavatsalam2020genericskb}, Ascent/Ascent++~\cite{ascent,ascentpp}, ATOMIC~\cite{atomic}, ASER~\cite{zhang2020aser} and TransOMCS~\cite{transomcs} do not have their focus on cultural knowledge and contain very little to zero assertions for geography or religion subjects, hence, they are not qualified for this comparison.

We evaluate 2 versions of \systemname{}, one where each base assertion is retained independently (\systemname{}-base-sent), the other containing only the cluster representatives (\systemname{}-cluster-reps).



 
 
 


\subsubsection{Setup}
For comparability, all resources are compared on 100 random assertions of the same 5 country subjects covered in StereoKG~\cite{stereokg} - United States, China, India, Germany and France.
We note that among all compared resources, Acharya et al.~\cite{acharya2020towards} only contain two subjects (United States and India), so for that resource, we only sample from those.
%
For StereoKG, we use their natural-language assertions. For Quasimodo and Acharya et al., we verbalize their triples using crafted rules.
Each assertion is evaluated by 3 MTurk annotators.
Additionally, we ask if the annotator would consider the assertion as an inappropriate or offensive material.
More details on the annotation task can be found in Appx.~\ref{apd:human-eval}.

\subsubsection{Results}
A summary of comparison with other resources is shown in Table~\ref{tab:kb-comparison}.

\paragraph{Resource size and assertion length}
\systemname{} outperforms all other resources on the number of base sentences. When turning to clusters, our resource still has significantly more assertions than Acharya et al. (which was constructed manually at small scale) and StereoKG (extracted from Reddit/Twitter questions). Quasimodo has comparable size with \systemname{}-cluster-reps for the country and religion domains and has more for the occupation domain.

The OpenIE-based methods, Quasimodo and StereoKG, produce the shortest assertion (32 and 37 characters on average, respectively). The manually-constructed KG (Acharya et al.) has the longest assertions (102 characters). \systemname{}, having average assertion lengths (69 and 73), stands between those two approaches.

\paragraph{Assertion quality}
In general, \systemname{}-cluster-reps considerably outperforms all other baselines on 2 of the 3 metrics (\textit{plausibility} and \textit{distinctiveness}). Our resource only comes behind Acharya et al. on the \textit{commonality} metric (1.15 and 1.22 respectively), which is expected because Acharya et al. only cover a few relations about common rituals (e.g., birthday, wedding)
in 2 countries, USA and India, and their assertions are naturally known by many workers on Amazon MTurk, who are mostly from these 2 countries~\cite{10.1145/1753846.1753873}. Importantly, the resource of Acharya et al. is based on crowdsourcing and only contains a small set of 225 assertions for a few rituals.


\systemname{}-cluster-reps even outperforms the manually-constructed KG (Acharya et al.) on the \textit{plausibility} metric. This could be caused by an annotation task design that is geared towards abnormalities, or lack of annotation quality assurance.

\systemname{} also has the highest scores on the \textit{distinctiveness} metric, while most of the assertions in other resources were marked as not distinguishing by the annotators.

Between the two versions of \systemname{}, the cluster representatives consistently outperform the base sentences on all evaluated metrics. This indicates that still some of the raw sentences in the collection are 
noisy,
on the other hand, the computed cluster representatives are more coherent and generally of better quality.

We also measured the \textit{offensiveness} (OFF) of each resource, i.e., the percentage of assertions that were marked as inappropriate or offensive materials by at least one of the human-annotators. Quasimodo and StereoKG, extracted from raw social media contents, have the highest number of assertions considered offensive (18\% and 13\%). Meanwhile, \systemname{}'s judicious filters only miss a small fraction (1\% of final assertions).

In summary, our \systemname{} CCSK collection has the highest quality by a large margin compared to other resources. Our resource provides assertions of high plausibility and distinctiveness. The clustering and cluster summarization also help to improve the presentation quality of the CCSK.

\begin{table}[t]
    \centering
    \caption{Assertion quality - \systemname{} vs. GPT-3 - evaluated on assertions of 196 countries. 
    }
    \label{tab:vs-gpt3}
    \small
    \begin{tabular}{lccccc}
    \toprule
    \multirow{2}{*}{\textbf{Method}} & \multicolumn{3}{c}{\textbf{Quality {[}0..2{]}}} & \multirow{2}{*}{{\begin{tabular}[c]{@{}r@{}}\textbf{OFF} \\ \textbf{(\%)}\end{tabular}}}  & \multirow{2}{*}{\textbf{LEN}} \\
    \cmidrule(lr){2-4}
    & \textbf{PLA} & \textbf{COM} & \textbf{DIS} \\
    \midrule
    \textbf{GPT-3}~\cite{GPT3} & 1.26 &	0.80 &	0.73 & 1 & 81 \\
    \rowcolor{gray!25}
    \textbf{\systemname{}} &	1.25 &	\textbf{0.89} &	\textbf{0.89} & 1 &	75 \\
    \bottomrule
    \end{tabular}
\end{table}

\begin{table*}[t]
    \centering
    \caption{Example assertions of \systemname{} and GPT-resource for subject:China, facet:clothing.}
    \label{tab:examples-lm}
    \footnotesize
    \begin{tabular}{l|p{.51\textwidth}|p{.435\textwidth}}
        \toprule
        \textbf{\#} & \textbf{\systemname{}} & \textbf{GPT-resource} \\
        \midrule
        1 & The bride usually wears red in a traditional Chinese wedding. &  Chinese people also like to wear modern clothes such as jeans and t-shirts. \\
        \midrule
        2 & The Chinese wear white at funerals bec.\ it is associated with mourning in Chinese culture. &  Shoes are also very important in Chinese culture.  \\
        \midrule
        3 & The Chinese wear new clothes for the New Year to symbolize new beginnings. &  Chinese people also like to dress their children in very cute clothes. \\
        \midrule
        4 & The costumes in Chinese opera are very colorful and important. & In China, you will often see little girls wearing dresses and boys wearing shorts. \\
        \midrule
        5 & In ancient China, only the emperor was allowed to wear the color yellow. & In the winter, people in China wear coats and scarves to keep warm. \\
        \bottomrule
    \end{tabular}
\end{table*}

\subsection{Comparison with direct LM extraction}\label{sec:comparison-lm}

Knowledge extraction directly from pre-trained LMs is recently popular, e.g., the LAMA probe~\cite{petroni2019language} or AutoTOMIC~\cite{west2021symbolic}.
There are major pragmatic challenges to this approach, in particular, that assertions cannot be contextualized with truly observed surrounding sentences, and that errors cannot be traced back to specific sources. 
Nonetheless, it is intrinsically interesting to compare assertion quality between extractive and generative approaches. In this section, we compare \systemname{} with assertions generated by the state-of-the-art LM, GPT-3~\cite{GPT3}.

 



\paragraph{Generating knowledge with GPT-3}
We query the largest GPT-3 model (\textit{davinci-002}) with the following prompt template:
{\em ``Please write 20 short sentences about notable <facet> in
<subject>.''}
We run each prompt 10 times and set the randomness (temperature) to 0.7, so as to obtain a larger resource.
We run the query for 5 facets and 210 subjects (196 countries and 14 religions), resulting in 188,061 unique sentences. Henceforth we call this dataset \emph{GPT-resource}, and reuse it in the extrinsic use cases (Sec.~\ref{sec:extrinsic}).

\paragraph{Evaluation metrics and setup}
For each resource, we sample 100 assertions for each facet (hence, 500 assertions in total) and perform human evaluation on the 3 metrics - commonality (COM), plausibility (PLA) and distinctiveness (DIS).

\paragraph{Results}
The quality comparison between assertions of \systemname{} and \textit{GPT-resource} is shown in Table~\ref{tab:vs-gpt3}.
While plausibility scores are the same, and \systemname{} performs better in commonality, the difference that stands out is in distinctiveness: GPT-3  performs significantly worse, reconfirming a known problem of language models, evasiveness and over-generality \cite{DBLP:journals/corr/LiGBGD15}. We illustrate this with anecdotal evidence in Table~\ref{tab:examples-lm}, for subject:China and facet:clothing.
None of the listed GPT-3 examples is specific for China.

\subsection{Extrinsic evaluation}\label{sec:extrinsic}

\paragraph{QA with context-augmented LMs}
Augmenting LMs input with additional contexts retrieved from knowledge bases has been a popular approach to question answering (QA) \cite{guu2020realm,petroni2020context}, which shows that although LMs store information in billions of parameters, they still lack knowledge to answer knowledge-intensive questions, e.g., \textit{``What is the appropriate color to wear at a Hindu funeral?''}

We use GPT-3 as QA agent, and compare its performance in 3 settings: (1) when only the questions are given, and when questions and their related contexts retrieved from (2) \systemname{} or (3) \textit{GPT-resource} (cf.\ Sec.~\ref{sec:comparison-lm}) are given to the LM. For \textit{questions}, we collect cultural knowledge quizzes from multiple websites, which results in 500 multiple-choice questions, each with 2-5 answer options (only one of them is correct). 
For \textit{context retrieval}, we use the the SentenceBert \textit{all-mpnet-base-v2} model, and for each question, retrieve the two most similar assertions from \systemname{}-cluster-reps and GPT-resource.
%
We use the GPT-3 \textit{davinci-002} model, 
with temperature=0 and max\_length=16
(see Appx.~\ref{apd:gpt-prompt} for prompt settings).

We measure the precision of the answers 
and 
present the results in Table~\ref{tab:qa}. 
It can be seen that with \systemname{} context, the performance is consistently better than when no context is given on all facets of culture, and better than GPT context on 3 out of 4 facets. 
This shows that GPT-3, despite its hundred billions of parameters, still lacks socio-cultural knowledge for question answering, and external resources such as \systemname{} CCSK can help to alleviate this problem.

\begin{table}[t]
\centering
\caption{Results of QA using context-augmented LMs.}
\label{tab:qa}
\small
\begin{tabular}{lcccc}
\toprule
\multirow{2}{*}{\textbf{Facet}} & \multirow{2}{*}{{\begin{tabular}[c]{@{}c@{}}\textbf{\#Ques-} \\ \textbf{tions}\end{tabular}}} & \multicolumn{3}{c}{\textbf{Precision (\%)}} \\
\cmidrule(lr){3-5}
& & \textbf{No cont.} & \textbf{GPT cont.} & \textbf{\systemname{} cont.} \\
\midrule
\textbf{Food/Drinks} & 88 & \textbf{92.05} & \textbf{94.32} & \textbf{93.18} \\
\textbf{Behaviors} & 125 & 60.80 & 57.60 & \textbf{63.20} \\
\textbf{Rituals} & 135 & 87.41 & 85.93 & \textbf{92.59} \\
\textbf{Traditions} & 152 & 72.37 & 69.74 & \textbf{79.61} \\
\midrule
\textbf{All} & 500 & 77.00 & 75.40 & \textbf{81.40} \\
\bottomrule
\end{tabular}
\end{table}

\paragraph{``Guess the country'' game}
The rule of this game is as follows: Given 5 CCSK assertions about a country, a player has to guess the name of the country.

As \textit{input}, we select a random set of 100 countries, and take assertions from either \systemname{} or \textit{GPT-resource}. 
The game has 5 rounds, each is associated with a facet of culture.
In each round, for each country, we draw the top-5 assertions from each resource (sorted by interestingness in \systemname{} or by frequency in \textit{GPT-resource}).
All mentions of the countries in the input sentences 
are replaced with \texttt{[...]}, before being revealed to the player.

This is a game that requires a player that possesses a wide range of knowledge across many cultures. Instead of human players, we choose GPT-3
as our player,
which has been shown to be excellent at many QA tasks~\cite{GPT3} (prompt settings are presented in Appx.~\ref{apd:gpt-prompt}).
%
%
% ...
%

We measure the precision of the answers
and present the
results 
in Table~\ref{tab:guess-game}. It can be seen that the player got significantly more correct answers when given assertions from \systemname{} than from \textit{GPT-resource} (i.e., assertions written by the player itself!). This confirms that assertions in \systemname{} are more informative.

\begin{table}[t]
\centering
\caption{Precision (\%) for the ``guess the country'' game.}
\label{tab:guess-game}
\small
\begin{tabular}{lcccccc}
\toprule
 & \multirow{2}{*}{\textbf{Food}} & \multirow{2}{*}{\textbf{Drinks}} & \multirow{2}{*}{{\begin{tabular}[c]{@{}r@{}}\textbf{Clo-} \\ \textbf{thing}\end{tabular}}} & \multirow{2}{*}{\textbf{Rituals}} & \multirow{2}{*}{{\begin{tabular}[c]{@{}r@{}}\textbf{Trad-} \\ \textbf{itions}\end{tabular}}} & \cellcolor{gray!25} \\
 & & & & & & \cellcolor{gray!25}\multirow{-2}{*}{\textbf{Avg.}} \\
\midrule
\textbf{GPT-resource} & 63.0 & 30.0 & 44.0 & 70.0 & \textbf{84.0} & \cellcolor{gray!25}58.2  \\
\textbf{\systemname{}} & \textbf{85.0} & \textbf{74.0} & \textbf{62.0} & \textbf{76.0} & 80.0 & \cellcolor{gray!25}\textbf{75.4} \\
\bottomrule
\end{tabular}
\end{table}



\section{Conclusion}

We presented \systemname{}---an end-to-end methodology for automatically collecting cultural commonsense knowledge (CCSK) from broad web contents at scale. We executed \systemname{} on several cultural subjects and facets of culture and produce CCSK of high quality. Our experiments showed the superiority of the resulting CCSK collection over existing resources, which have limited coverage for this kind of knowledge, and also over methods based on prompting LMs. 
Our work expands CSKG construction into a domain that has been largely ignored so far.
Our data and code are accessible at \website{}.




\subsection*{Ethics statement}
No personal data was processed and hence no IRB review was conducted. It is in the nature of this research, however, that some outputs reflect prejudices or are even offensive. We have implemented multiple filtering steps to mitigate this, and significantly reduced the percentage of offensive assertions, compared with prior work. Nonetheless, \systemname{} represents a research prototype, and outputs should not be used in downstream tasks without further thorough review.


\bibliographystyle{ACM-Reference-Format}
\bibliography{refs}

\appendix

\section{System summary}\label{apd:system-summary}

We provide an overview of techniques and models applied in each module of the \systemname{} system in Table~\ref{tab:list-techniques}.

\begin{table}[h]
    \centering
    \footnotesize
    \caption{List of models and techniques used in each module of \systemname{}.}
    \begin{tabular}{cp{.3\columnwidth}p{.55\columnwidth}}
    \toprule
        \textbf{\#} & \textbf{Module} & \textbf{Used techniques/models} \\
    \midrule
         1 & Subject detection (Sec.~\ref{sec:subject-detection}) & String matching, NER \\
    \midrule
         2 & Generic assertion filtering (Sec.~\ref{sec:generic-filtering}) & Hand-crafted lexico-syntactic rules \\
    \midrule
         3 & Cultural facet classification (Sec.~\ref{sec:classification}) & Huggingface's \texttt{bart-large-mnli} model for the textual entailment task \\
    \midrule
         4 & Assertion clustering (Sec.~\ref{sec:clustering}) & SentenceBert and HAC algorithm for clustering, GPT-3 for cluster summarization \\
    \midrule
         5 & Concept extraction (Sec.~\ref{sec:concept-extractino}) & Common n-gram extraction \\
    \midrule
         6 & Cluster ranking + post-filtering (Sec.~\ref{sec:ranking}) & Ad-hoc ranking features, rule-based filtering \\
    \bottomrule
    \end{tabular}
    \label{tab:list-techniques}
\end{table}

\section{Generic filtering rules}\label{apd:filtering}
GenericsKB~\cite{bhakthavatsalam2020genericskb} was built by using a set of 27 hand-crafted lexico-syntactic rules to extract high-quality generic sentences from different text corpora (the ARC corpus, SimpleWikipedia and the Waterloo crawl of education websites). For example, the lexical rules look for sentences with short length, starting with a capitalized character, having no bad first words (e.g., determiners), ending with a period, having no URL-like snippets, etc. The syntactic rules only accept a sentence if its root is a verb and not the first word, and if there is a noun before the root verb, etc. 

\systemname{} adopts the GenericsKB rules. However, as GenericsKB only deals with general concepts (e.g., ``tree'', ``bird'', etc.), some of the rules are not applicable for the cultural subjects that can be named entities. Hence, depending on the subjects and facets, we adaptively modify the rules (by dropping some of them) so that we will not miss out valuable assertions. For instance, for geography, the \textit{has-no-determiners-as-first-word} rule will filter out valuable assertions such as \stmt{The Chinese use chopsticks to eat their food} or \stmt{The currywurst is a traditional German fast food dish}, and it must be dropped. In another situation, when exploring the ``traditions'' facet, the \textit{remove-past-tense-verb-roots} rule would be too aggressive as it rejects assertions about past traditions. 
The rule that rejects sentences with PERSON entities can be used for the geography and occupation subjects, but must not be used for religions, because it will filter out sentences about Buddha or Jesus Christ. Full details are in the published code base\footnote{\url{https://github.com/cultural-csk/candle/blob/main/candle/pipeline/component_generic_sentence_filter.py}}.

\section{Hyperparameter settings}\label{apd:hyperparameters}
Based on tuning on small withheld data, we select the following values for hyperparameters and run \systemname{} on the C4 dataset with these settings.
For \textit{cultural facet classification} (cf. Sec.~\ref{sec:classification} and Eq.~\ref{eq:classification}), we fix $\rho_+$ to 0.5 and $\rho_-$ to 0.3. 
%
For \textit{assertion clustering} (cf. Sec.~\ref{sec:clustering}), we use the SentenceBert model \textit{all-MiniLM-L6-v2} for computing sentence embeddings. For the HAC algortihm, we measure point-wise Euclidean distance of the normalized embeddings. Then, we use the Ward's linkage~\cite{Ward1963HierarchicalGT}, with the maximal distance threshold set to 1.5. 
In the few cases where input sets are larger, we truncate them at 50K sentences per subject-facet pair, since larger inputs only contain further redundancies, that are not worth the cubic effort of clustering. This concerns only 15 out of 386 subjects.
%
For \textit{cluster summarization}, 
we 
consider the 500 most populated clusters for each subject-facet pair with a minimum size of 3 sentences. 
%
For \textit{cluster ranking} (cf. Sec.~\ref{sec:ranking}), we fix $\theta$ in Eq.~\ref{eq:distinctiveness} to 0.8.



\section{Intrinsic evaluation}\label{apd:intrinsic}
%

We break down the \systemname{} CCSK collection into domains and facets and evaluate the assertion quality for each of these sub-collections and get more insights into the produced data.

\begin{table}[t]
\centering
\caption{Quality of \systemname{} assertions for each domain.}
\label{tab:socco-domains}
\footnotesize
\begin{tabular}{lcccccc}
\toprule
\multirow{2}{*}{\textbf{Domain}} & \multicolumn{3}{c}{\textbf{Quality {[}0..2{]}}} & \multicolumn{3}{c}{\textbf{Acceptance rate (\%)}} \\
\cmidrule(lr){2-4} \cmidrule(lr){5-7}
& \textbf{PLA} & \textbf{COM} & \textbf{DIS} & \textbf{PLA $\ge$ 1} & \textbf{COM $\ge$ 1} & \textbf{DIS $\ge$ 1} \\
\midrule
\textbf{Geography} & 1.52 & 1.19 & 1.03 & 84 .00& 66.00 & 61.33 \\
\textbf{Religion} & 1.51 & 1.29 & 1.22 & 85.76 & 74.67 & 72.00 \\
\textbf{Occupation} & 1.59 & 1.50 & 1.25 & 86.67 & 82.67 & 73.67 \\
\midrule
\textbf{Average} & 1.54 & 1.33 & 1.17 & 85.44 & 74.44 & 69.00 \\
\bottomrule
\end{tabular}
\end{table}

\subsection{Per-domain quality}\label{subsec:per-domain-quality}
\systemname{} contains 3 cultural domains - geography, religion and occupation. For each domain, we sample 100 assertions and perform crowdsourcing evaluation with the 3 metrics - PLA, COM and DIS (cf. SubSec.~\ref{subsec:metrics}). 
We present the evaluation results in Table~\ref{tab:socco-domains}. 
Besides the raw scores (0, 1, 2), we also binarize and denote them as acceptance rates, i.e., a score greater than zero means \textit{``accept''}.

\systemname{} achieves a high \textit{plausibility} (PLA) score of 1.54 on average. Performance on this metric is relatively consistent through all domains. 
Meanwhile, the \textit{commonality} (COM) metric is highest for the occupation domain and lowest for geography domain. 

More than 80\% of plausible assertions are annotated as \textit{distinctive} (DIS). Religion and occupation assertions perform significantly better than geography's on this metric. That could be caused by several assertions for geography subjects being correct but too generic (e.g., \stmt{Japanese food is enjoyed by many people}, or \stmt{German beer is good}). On the other hand, religions and occupations are more distinguishing from one another, while countries or geo-regions usually have cultural overlaps.
 
\begin{table}[t]
\centering
\caption{Quality of \systemname{} assertions for each facet and the domain \textit{geography>country}.}
\label{tab:socco-countries}
\footnotesize
\begin{tabular}{lcccc}
\toprule
\multirow{2}{*}{\textbf{Facet}} & \multicolumn{4}{c}{\textbf{Quality {[}0..2{]}}} \\
\cmidrule(lr){2-5}
 & \multicolumn{1}{l}{\textbf{DOM}} & \multicolumn{1}{l}{\textbf{PLA}} & \multicolumn{1}{l}{\textbf{COM}} & \multicolumn{1}{l}{\textbf{DIS}} \\
\midrule
\textbf{Food} & 1.42 & 1.23 & 0.94 & 0.97 \\
\textbf{Drinks} & 1.51 & 1.40 & 1.14 & 1.19 \\
\textbf{Clothing} & 1.49 & 1.30 & 1.04 & 1.07 \\
\textbf{Rituals} & 1.45 & 1.27 & 1.06 & 1.20 \\
\textbf{Traditions} & 1.42 & 1.27 & 1.02 & 1.11 \\
\midrule
\textbf{Average} & 1.46 & 1.29 & 1.04 & 1.11 \\
\bottomrule
\end{tabular}
\end{table}

\subsection{Per-facet quality}\label{subsec:per-facet-quality}
We select the assertions for the domain country, and for each facet (food, drinks, clothing, traditions, rituals) we sample 100 assertions for crowdsourcing evaluation. Besides commonality (COM), plausibility (PLA) and distinctiveness (DIS), here we introduce one more evaluation metric, domain relevance (DOM), to measure if an assertion talks about the cultural facet of interest. Only when the DOM score 
is greater than zero, the other metrics will be evaluated. We present the evaluation results in Table~\ref{tab:socco-countries}.

It can be seen that \systemname{} maintains good quality on all evaluation metrics. 
Notably, scores for the DOM metric are consistently high for all facets, suggesting that the enhanced techniques for zero-shot classification work well on our data.
Interestingly, the facet \textit{drinks} outperforms all other facets on 3 of the 4 metrics (DOM, PLA and COM), especially for PLA, its score is significantly higher than others.
Assertions for \textit{drinks} and \textit{rituals} are also more distinctive than for other facets.

\section{Details of Annotation Task for Assertion Evaluation}\label{apd:human-eval}
The evaluations of assertion quality (Tables \ref{tab:kb-comparison}, \ref{tab:vs-gpt3}, \ref{tab:socco-domains} and \ref{tab:socco-countries}) are conducted on Amazon MTurk (AMT). We present CCSK assertions to annotators in the form of natural-language sentences (triples from Quasimodo~\cite{quasimodo} and Acharya et al.~\cite{acharya2020towards} were verbalized using crafted rules).
We evaluate each assertion along 3-4 dimensions on a 3-point Liker scale - negation (0), ambiguity (1) and affirmation (2). 
Each AMT task consists of 5 assertions evaluated by 3 different annotators. Workers are compensated \$0.50 per task. We select Master workers with lifetime's acceptance rate more than 99\%. We obtain fair inter-annotator agreements given by Fleiss' kappa~\cite{fleiss1973equivalence}: 25.0 for DOM, 25.7 for PLA and 25.4 for DIS. This number for COM (13.4) is lower than others because it is an objective question (has the annotator heard of the assertion?).

\section{GPT-3 prompting}\label{apd:gpt-prompt}
In this work, we use GPT-3 for cluster summarization (Sec.~\ref{sec:clustering}), generating CCSK for \textit{GPT-resource} (Sec.~\ref{sec:comparison-lm}), context-augmented QA and ``guess the country'' game (Sec.~\ref{sec:extrinsic}). The prompt templates and settings used for these tasks are presented below.

\begin{figure}[t]
    \centering
    \includegraphics[width=.8\columnwidth]{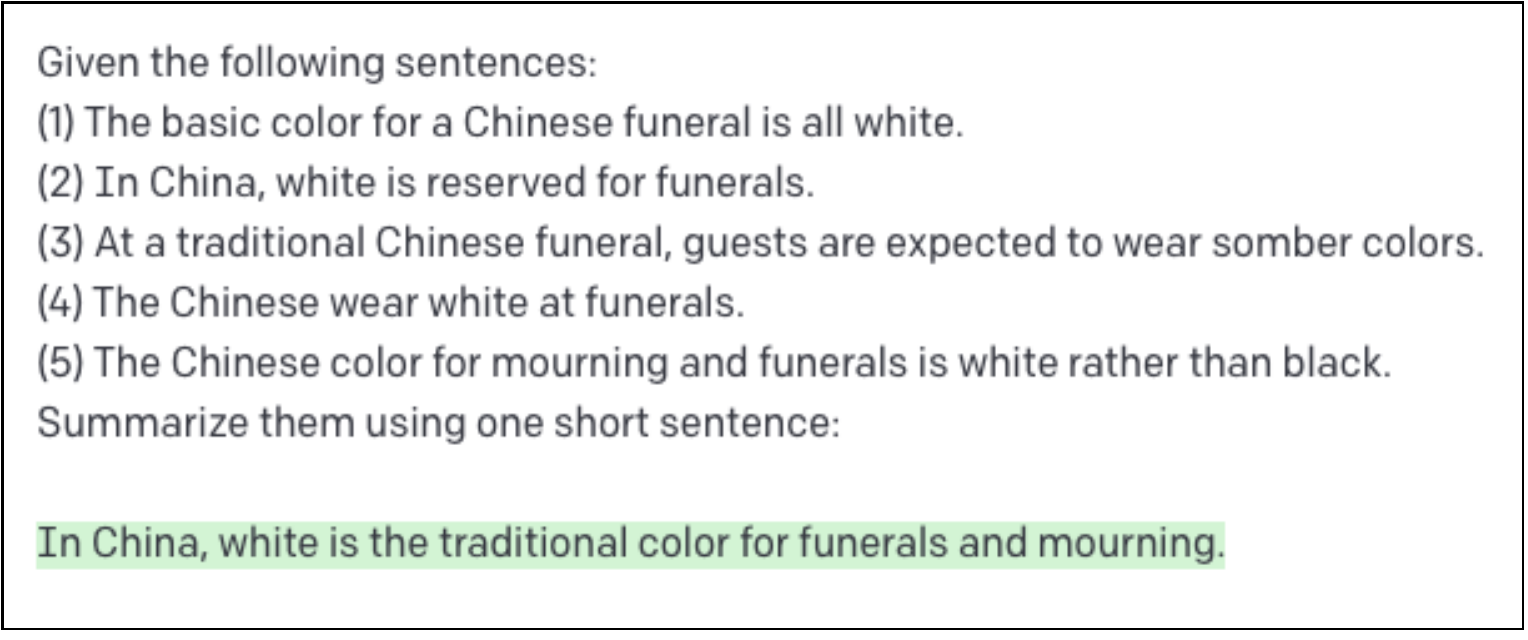}
    \caption{A screenshot of GPT-3 output for cluster summarization.}
    \label{fig:summarize-prompt}
\end{figure}

\paragraph{Cluster summarization}
We query the \textit{curie-001} model, with zero temperature and maximum length of 50 tokens.
We only take the first generated sentence as output.
An example prompt is presented in Fig.~\ref{fig:summarize-prompt}.

\paragraph{Generating CCSK for GPT-resource}
We use the \textit{davinci-002} model and set temperature to 0.7 and maximum length to 512 tokens. For each facet and subject, we run the following prompt template for 10 times: \texttt{\small Please write 20 short sentences about notable <facet> in
<subject>.} We query for 5 facets (food culture, drinking culture, clothing habits, rituals, traditions), and 210 subjects (196 countries and 14 religions).
Table~\ref{tab:examples-lm}
shows
some example generations
for the subject China and the facet ``clothing habits''.

\begin{figure}[t]
    \centering
    \includegraphics[width=\columnwidth]{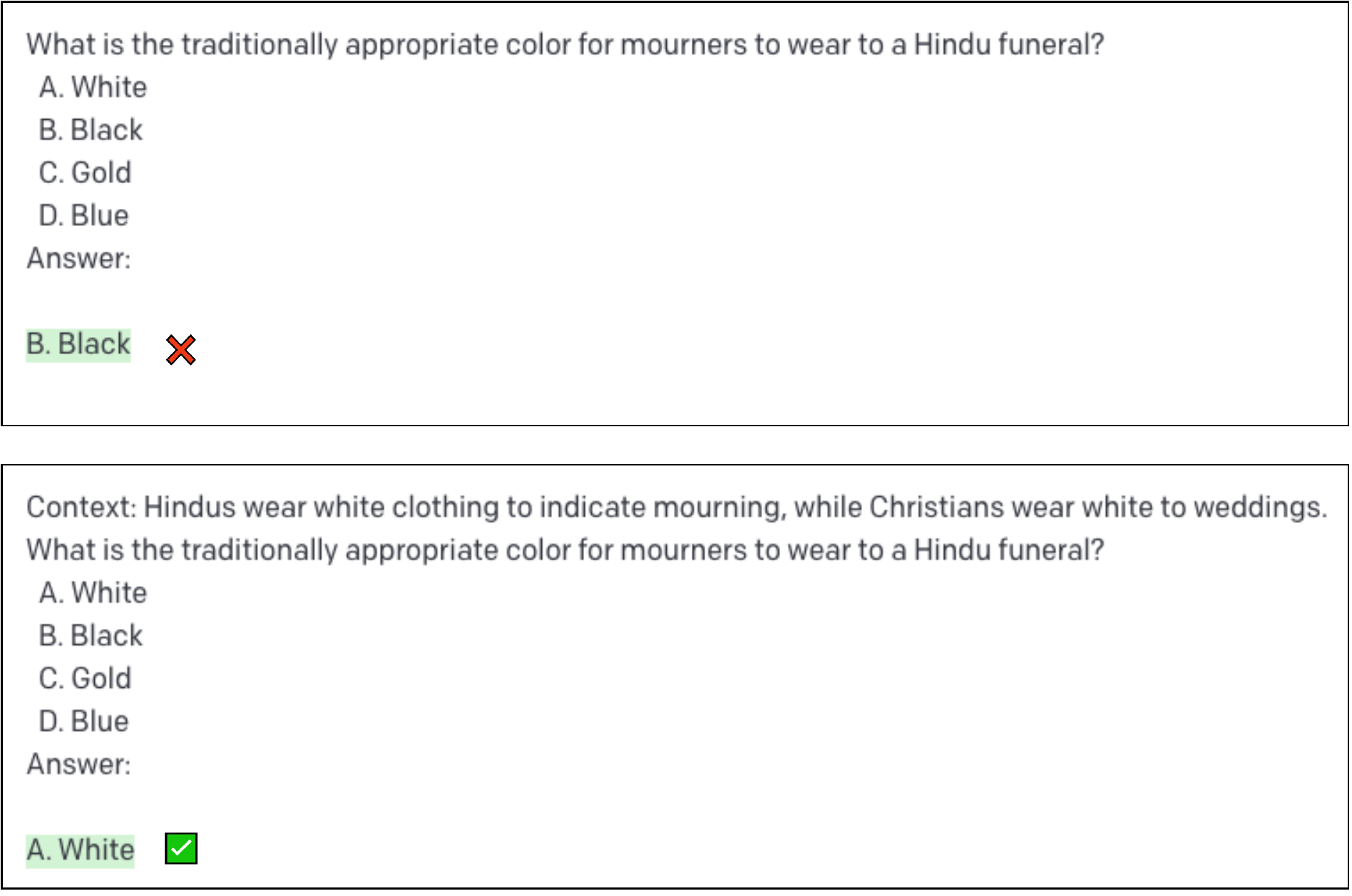}
    \caption{Screenshots of GPT-3 output in the QA task, without and with CCSK.}
    \label{fig:qa-prompt}
\end{figure}

\paragraph{Context-augmented QA}
We query the \textit{davinci-002} model with zero temperature and maximum length of 16 tokens. Answers are then manually mapped to the respective options. Example prompts are shown in Fig.~\ref{fig:qa-prompt}.

\begin{figure}[t]
    \centering
    \includegraphics[width=\columnwidth]{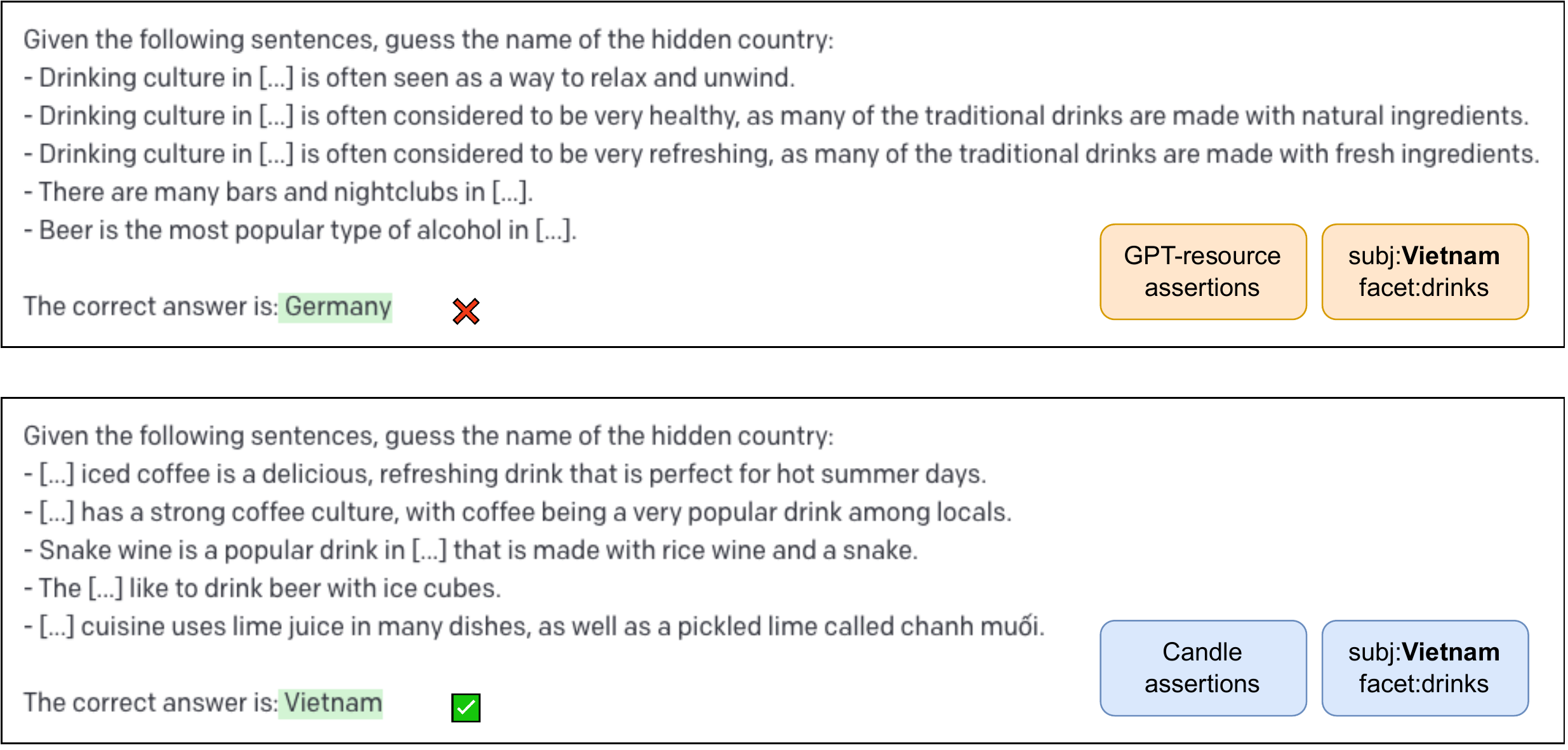}
    \caption{Screenshots of GPT-3 output for the ``guess the country'' game, with assertions of \textit{GPT-resource} and \systemname{} for subject:\textit{Vietnam} and facet:\textit{drinks}.}
    \label{fig:guess-country-prompt}
\end{figure}

\paragraph{``Guess the country'' game}
We use the \textit{davinci-002} model, with temperature=0 and a max\_length=8. 
Answers given by GPT-3 are checked manually for their correctness. Example prompts can be seen in Fig.~\ref{fig:guess-country-prompt}.

\end{document}